\documentclass[runningheads]{llncs}

\usepackage{amssymb}
\usepackage{graphicx}

\usepackage{url}
\usepackage{color}
\usepackage{amsmath}
\usepackage{amsfonts}
\usepackage{blindtext}

\usepackage{mathrsfs}
\usepackage{bbm}
\usepackage{url}
\usepackage{algorithm}
\usepackage{algorithmic}
\usepackage{paralist}
\usepackage[english]{babel}
\usepackage{booktabs}

\usepackage{multicol}
\usepackage{multirow}
\usepackage{subfig}

\begin{document}
\title{Automatic Short Answer Grading via Multiway Attention Networks}
%
%
\author{Tiaoqiao Liu\inst{1} \and
Wenbiao Ding\inst{1} \and
Zhiwei Wang\thanks{Work was done when the authors did internship in TAL AI Lab}\inst{2} \and
Jiliang Tang\inst{2} \and Gale Yan Huang\inst{1} \and Zitao Liu\thanks{Corresponding Author: Zitao Liu}\inst{1}}

\authorrunning{T. Liu et al.}
%

\institute{TAL AI Lab, Beijing, China  \\
\email{\{liutianqiao,dingwenbiao,galehuang,liuzitao\}@100tal.com}
\and Data Science and Engineering Lab, Michigan State University, USA\\
\email{\{wangzh65,tangjili\}@msu.edu}}
\maketitle              

\begin{abstract}
Automatic short answer grading (ASAG), which autonomously score student answers according to reference answers,  provides a cost-effective and consistent approach to teaching professionals and can reduce their monotonous and tedious grading workloads. However, ASAG is a very challenging task due to two reasons: (1) student answers are made up of free text which requires a deep semantic understanding; and (2) the questions are usually open-ended and across many domains in K-12 scenarios. In this paper, we propose a generalized end-to-end ASAG learning framework which aims to (1) autonomously extract linguistic information from both student and reference answers; and (2) accurately model the semantic relations between free-text student and reference answers in open-ended domain. The proposed ASAG model is evaluated on a large real-world K-12 dataset and can outperform the state-of-the-art baselines in terms of various evaluation metrics. 

\end{abstract}

\section{Introduction}
\label{sec:intro}
Assessing the knowledge acquired by students is one of the most important aspects of the learning process as it provides feedback to help students correct their misunderstanding of knowledge and improves their overall learning performance. Traditionally, the assessing paradigm is often conducted by instructors or teachers. However, this access paradigm is not suitable in many cases especially when teaching resources are not readily available. To address this gap, many computer-assisted assessment approaches are developed to automate the assessment process \cite{daradoumis2013review}.

One specific task, automatic short answer grading (ASAG), whose objective is to automatically score the free-text answers from students according to the corresponding reference answer \cite{saha2018sentence}, has attracted great attentions from a variety of research communities and some promising results have been already obtained \cite{nielsen2009recognizing,ramachandran2015identifying,mitchell2002towards,saha2018sentence,sultan2016fast}. However, ASAG still remains challenging mainly for two reasons. Firstly, the student answers are expressed in different ways of free texts. Thus, it requires the ASAG approach to have a deep semantic understanding of the student answers. Secondly, the questions or assessments (and the corresponding reference answers) usually are open-ended and across different domains. The ASAG approach should be general and applicable into different scenarios. 

In this paper, to address challenges above, we take the advantage of recent advances in natural language processing field \cite{vaswani2017attention,devlin2018bert} and propose a deep learning framework to tackle the ASAG problem in an end-to-end approach. Specifically, our framework utilizes attention mechanisms to understand the semantics of student and reference answers with most relevant information and is very flexible and efficient as it can be easily extended with extra neuron layers while still maintaining fast training speed thanks to its high parallelization ability. Our main contributions are summarized as follows: (1) We propose an end-to-end approach that does not require any feature engineering effort to tackle the short answer grading problem; (2)We develop a novel framework that is able to modeling the relation between student and reference answers by accurately identifying matching information and understanding the semantic meaning; and (3) The proposed framework can be used in a wide range of domains and is easily scalable for large-scale datasets. It is demonstrated on a large-scale real-world dataset collected from millions of K-12 students.


\section{Our Approach}
\label{sec:method}

\begin{figure}[!tpbh]
\centering
\includegraphics[width=\textwidth] {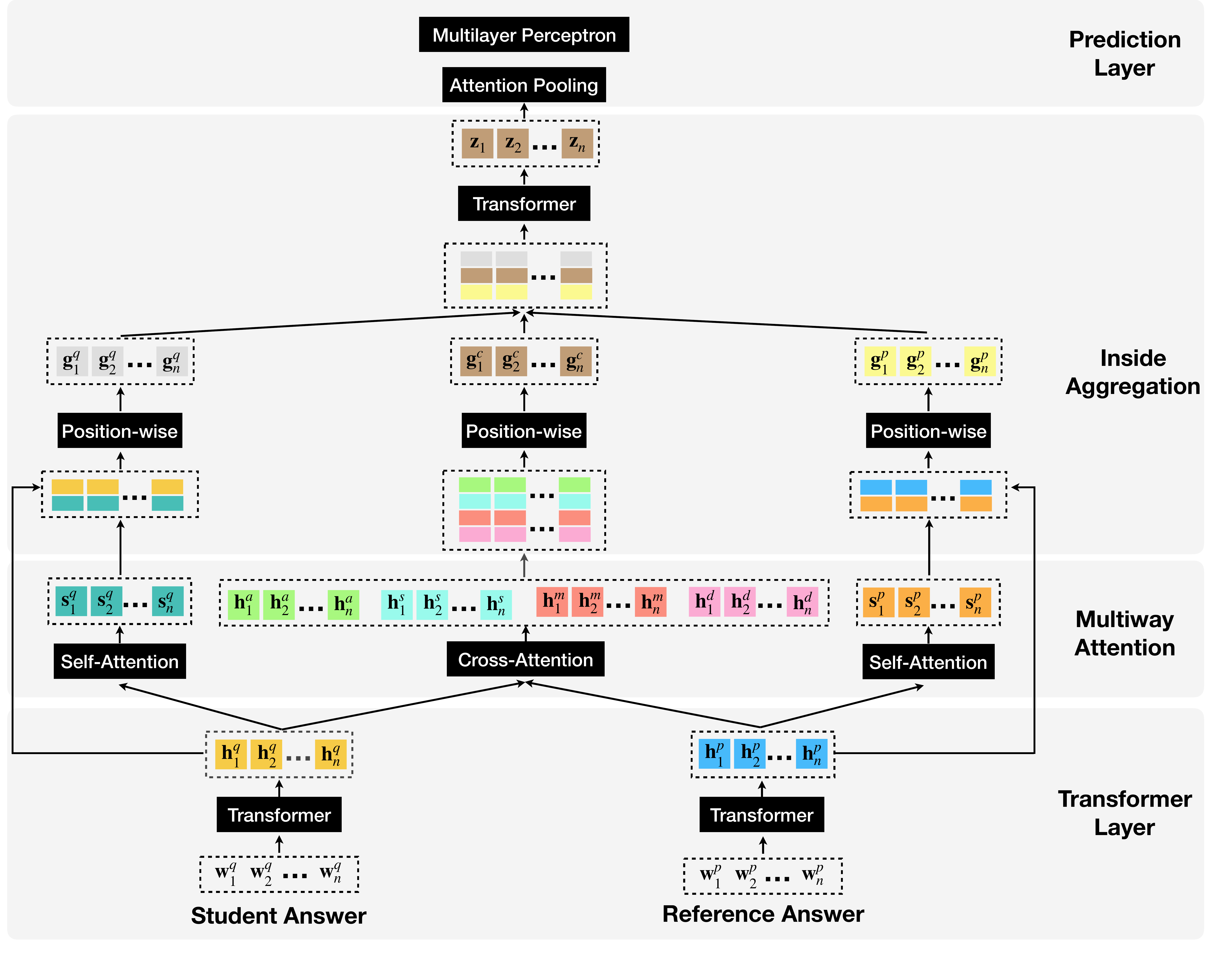}
\caption{The overview of our model (better viewed in color).}
\label{fig:model}
\vspace{-0.7cm}
\end{figure}

In  this  section,  we  introduce  our  proposed  framework, the overall structure is shown in Figure \ref{fig:model}. Before detailing each component next, we first introduce the notations. We use bold lower case letters for vectors and bold upper case letters for matrices. We use subscript to represent the vector index, which is the index of word in each sentence in most cases. We also use superscript to represent the category of vectors.

\noindent{\bf Transformer Layer.} The input of the transformer layer is the student and reference answer, which are two sequences of words and denoted as $\{{\bf w}_1^q, {\bf w}_2^q, \cdots, {\bf w}_n^q\}$ and $\{{\bf w}_1^p, {\bf w}_2^p, \cdots, {\bf w}_n^p\}$, respectively, where $\{\mathbf{w}_i^q\}$ and $\{\mathbf{w}_i^p\}$ are the pre-trained word embeddings. Next, the {\it transformer}~\cite{vaswani2017attention} model is applied as: $\{{\bf h}_1^*, {\bf h}_2^*, \cdots, {\bf h}_n^*\}$  = {\it transformer}(${\bf w}_1^*, {\bf w}_2^*, \cdots, {\bf w}_n^*)$, where $*\in \{p, q\}$ and  each $\{\mathbf{h}_i^q\}$ and  $\{\mathbf{h}_i^p\}$ are the word embeddings that contain its contextual sentence information in the student and reference answers, respectively.  


\noindent{\bf Multiway Attention.} We design the multiway attention layer to capture the relations between student and reference answers. Specifically, it consists of two blocks. The first is self-attention block where each ${\bf h}_i^*$ will attend every ${\bf h}_j^*, j \in \{1,2,\cdots, n\}$ to obtain new representation ${\bf s}_i^*$, $*\in \{p, q\}$. The second is cross-attention block in which each ${\bf h}_i^q$ will attend every ${\bf h}_j^p, j \in \{1,2,\cdots, n\}$ to obtain another set of new representations ${\bf h}_i^t, t \in \{a, s, m, d\}$, where $a, s, m, d$ are addictive, subtractive, multiplicative, and dot-product attention mechanisms, respectively~\cite{tan2018multiway}.

\noindent{\bf Inside Aggregation.} This layer is designed to aggregate multiway attention layer outputs to a single representation ${\bf z}$. Specifically, we first concatenate the outputs from cross-attention and self-attention blocks by positions respectively and feed them to different position-wise feed forward networks to obtain the compressed representations ${\bf g}_i^*$, $*\in \{p, q, c\}$, where $p, q, c$ represent student answer sequence, reference answer sequence, and cross-attention sequence, respectively. We concatenate the outputs ${\bf g}_i^*$ by positions and  after another Transformer block, we get new sequence representation ${\bf Z} = transformer([g_i^p, g_i^q, g_i^c]), i \in \{1,2,\cdots, n\}$ which contains the information in student and reference answers and the relations between them.

\noindent{\bf Prediction Layer.} The evaluation of student answer will be produced by this layer. Specifically, we first convert the aggregated sequence representation ${\bf Z}$ to a fixed-length vector with self-attention pooling layer. This transformation is defined as: ${\bf x} = softmax({\bf w_1^z} tanh({\bf W_2^z} {\bf Z^T})){\bf Z}$, where ${\bf w_1^z}$ and ${\bf W_2^z}$ are learned parameters during training step. Then we build a feed forward network that takes ${\bf x}$ as input and outputs a two-dimensional vector. The output vector is sent to a softmax function to obtain the final probabilistic evaluation vector. The first entry gives the probability of wrong answer while the second entry gives right answer probability. The objective is to minimize the cross entropy of the relevance labels.

\section{Experiments}
\label{sec:experiment}
In this section, we conduct experiments on a large real-world educational data, which contains 120,000 pairs of student answers and question analysis from an online education platform, each labeled with binary value indicating whether the student has the right answer. The positive and negative instances are balanced and we randomly select 30,000 samples as our test data and use the rest for validation and training. The hyperparameters of our model are selected by internal cross validation. We use both AUC and accuracy as our evaluation metrics and for both metrics, a higher value indicates better performance. 

We compare our model with several state-of-the-art baselines. More specifically, we choose: (1) Logistic regression (LR). (2) Gradient boosted decision tree (GBDT) \cite{Friedman1999StochasticGB,Ye2009StochasticGB}.  (3) Multichannel convolutional neural networks (TextCNN)~\cite{kim2014convolutional}.  (4) Sentence embedding by Bidirectional Transformer block (Bi-Transformer)~\cite{vaswani2017attention}. (5) Multiway Attention Network (MAN)~\cite{tan2018multiway}. and (6) Manhattan LSTM with max pooling (MaLSTM)~\cite{mueller2016siamese}.

\subsection{Experimental Results} 

We report the experimental results in Table \ref{tab:asag}. From the table, we observe that our model outperforms all of the baselines. We argue that this is because our model is able to effectively capture the semantic  information  between  student and reference  answers.  This is confirmed by the fact that MAN shows the superior performance among all baselines, as it not only aggregates sentence information within Transformer block, but matches words in both query sentence and answer sentence from multiple attention functions.

\begin{table}
\vspace{-0.4cm}
\centering
\caption{ASAG performance comparison on a real-world K-12 dataset.}
\label{tab:asag}
\begin{tabular}{l|l|l|l|l|l|l|l}
\toprule
 & LR & GBDT & TextCNN & Bi-Transformer & MaLSTM & MAN & Our\\ \midrule
\textbf{Accuracy} & 0.8297 & 0.8628 & 0.8772 & 0.8813 & 0.8825 & 0.8808 & {\bfseries 0.8899}\\ \midrule
\textbf{AUC} & 0.8808 & 0.9287 & 0.9312 & 0.9335 & 0.9375 & 0.9365 & {\bfseries 0.9444}\\
\bottomrule
\end{tabular}
\vspace{-0.4cm}
\end{table}

\section{Conclusion}
\label{sec:conclusion}
\vspace{-0.3cm}
In this paper we present our multi-way attention network for automatic short answer grading. We use transformer blocks and attention mechanisms to extract answer matching information. To comprehensively capture the semantic relations between the reference answer and the student answers, we apply multiway attention functions instead of single attention channel. Experiment results on a large real-world education dataset demonstrate the effectiveness of the proposed framework. There are several directions that need further exploration. We may use one attention mechanism with multiple heads instead of multiple attention mechanisms and we may replace transformer block with other type of sentence encoder like self-attention network or hierarchical attention network.

%
%
\bibliographystyle{splncs04.bst}
\bibliography{aied2019}
\end{document}